\documentclass[preprint]{article}

\usepackage{xspace}
\newcommand{\sys}{SchedCP\xspace}
\newcommand{\agent}{sched-agent\xspace}

\usepackage{neurips_2025}

\usepackage{listings}     
\usepackage{booktabs}     
\usepackage{array}        
\usepackage{tabularx}     
\usepackage{enumitem}     

\usepackage{comment}

\usepackage[utf8]{inputenc}
\usepackage[T1]{fontenc}
\usepackage{textcomp}
\usepackage[english]{babel} 
\usepackage{url}
\usepackage{graphicx}
\usepackage{subcaption}     

\usepackage{hyperref}       
\usepackage{amsfonts}       
\usepackage{nicefrac}       
\usepackage{microtype}      
\usepackage{xcolor}         

\title{Towards Agentic OS: An LLM Agent Framework for Linux Schedulers}

\author{%
  Yusheng Zheng$^{1}$ \quad
  Yanpeng Hu$^{2}$ \quad
  Wei Zhang$^{3}$ \quad
  Andi Quinn$^{1}$ \\
  $^{1}$UC Santa Cruz, CA, USA \quad
  $^{3}$University of Connecticut \quad
  $^{2}$ShanghaiTech University, Shanghai, China \\
  \texttt{\{yzhen165, aquinn1\}@ucsc.edu, huyp@shanghaitech.edu.cn}
}

\begin{document}

\maketitle

\begin{abstract}
Operating system schedulers suffer from a fundamental semantic gap, where kernel policies fail to understand application-specific needs, leading to suboptimal performance. We introduce \sys, the first framework that enables fully autonomous Large Language Model (LLM) agents to safely and efficiently optimize Linux schedulers without human involvement. Our core insight is that the challenge is not merely to \emph{apply} a better LLM, but to architect a decoupled control plane that separates the AI’s role of semantic reasoning ("what to optimize") from the system’s role of execution ("how to observe and act"), thereby separating the optimization problem into two stages: goal-inference and policy-synthesis. Implemented as Model Context Protocol(MCP) server, \sys provides a stable interface with three key services: a Workload Analysis Engine, an evolving Scheduler Policy Repository, and an Execution Verifier that validates all AI-generated code and configurations before deployment with static and dynamic analysis. We demonstrate this architecture's power with \agent, a multi-agent system that autonomously analyzes workloads, synthesizes custom eBPF scheduling policies, and deploys them via the sched\_ext infrastructure. Our evaluation shows that SchedCP achieves up to 1.79x performance improvement and 13x cost reduction compared to naive agentic approaches, all while maintaining high success rate. The code is in \href{https://github.com/eunomia-bpf/schedcp}{https://github.com/eunomia-bpf/schedcp}.
\end{abstract}

\maketitle
\section{Introduction}
\label{sec:intro}

Operating system schedulers face a fundamental challenge: kernel policies cannot understand what applications need, leading to suboptimal performance as Linux's EEVDF scheduler~\cite{eevdf2024} applies one-size-fits-all policies to diverse workloads. While sched\_ext~\cite{schedext2024} in Linux 6.12 enables custom extended Berkeley Packet Filter(eBPF) schedulers with safety guarantees through verification, developing them still requires both deep kernel expertise and a good understanding of the workloads.

Prior reinforcement learning-based schedulers~\cite{mao2019decima,qiu2020firm} lack semantic understanding of workloads, are often limited to tweaking configurations within a problem space predefined by human engineers, preventing fully automatic system optimization. While LLMs~\cite{openai2023gpt4,anthropic2024claude} and agent frameworks~\cite{autogen,geminicli,claudecode,qian2024chatdev,hong2023metagpt} excel at code generation, naively applying them to scheduler development proves impractical: our experiments show generating a basic scheduler takes 33 minutes, costs \$6, and often degrades performance. The gap remains: existing methods lack semantic understanding, while LLMs lack the scaffolding for safe, efficient, and reliable systems integration despite prior LLM-eBPF work~\cite{kgent}.

Our work enables a system to drive its own optimization by decomposing the problem into two stages: a goal-inference stage that uncovers optimization goals and constraints from workloads, and a policy-synthesis stage that compiles them into eBPF scheduler policies. This approach is embodied in a decoupled architecture with two components. First, \sys\ is a control plane framework providing safe AI-kernel interfaces with profiling, tracing and validation tools, allowing LLMs to customize the kernel scheduler with eBPF. It exposes kernel scheduling features via Model Context Protocol (MCP) through three core services: Workload Analysis Engine, Scheduler Policy Repository, and Execution Verifier. Second, \agent\ is the first autonomous multi-agent system that decomposes scheduler optimization into four specialized agents (Observation, Planning, Execution, Learning), demonstrating how LLMs can bridge the semantic gap between application requirements and kernel scheduling policies. This separation allows \sys\ to provide a generalizable framework for any AI agent, while \agent\ demonstrates semantic workload analysis and policy generation. Our evaluation shows \agent\ achieves up to 1.79× performance on kernel compilation, 2.11× P99 latency improvement and 1.60× throughput gain on schbench, 20\% latency reduction for batch workloads, and 13× cost reduction.

\section{Motivation}
\label{sec:motivation}

Linux scheduler optimization faces three barriers. First, a domain knowledge gap exists between developers and users: DevOps engineers lack insight into workload characteristics (latency-sensitive vs. throughput-oriented), while edge/personal device users lack both kernel optimization expertise and understanding of application-specific targets. Second, scheduler development requires mastering kernel programming, limiting innovation to a few experts. Third, modern workloads exhibit complex dynamics: web traffic varies by orders of magnitude daily, build system parallelism changes with dependencies. Prior RL-based schedulers~\cite{mao2019decima,qiu2020firm,zhang2024mrsch,mao2019park} require extensive training per workload type, lack semantic understanding to transfer across workloads, and only tweak configurations after engineers have already defined the entire problem space: selecting features, specifying knobs, and writing objective functions. LLMs uniquely bridge these gaps by: (1) using tools to dynamically explore diverse workloads to uncover application intent and structure, which is difficult to capture with conventional static or dynamic analysis; (2) applying a broad, pre-trained understanding of source code semantics to reason about performance trade-offs and structural dependencies, such as data-dependency hints; (3) synthesizing correct eBPF schedulers based on this analysis; and (4) operating in the control plane to generate optimized code that runs natively with negligible runtime overhead, unlike traditional ML models that would cause unacceptable inference latency in the scheduler hot path.

We tested Claude Code\cite{claudecode}, the  state-of-the-art LLM agent, with "write a FIFO scheduler in eBPF" from an empty folder, with all permissions and bash access. Of three attempts, only one succeeded. The second attempt produced pseudo-code after 6 minutes trying, and the third generated a scheduler tracer instead after 8 minutes of development. The successful generation required 33 minutes, 221 LLM API calls, and 15+ iterations, costing \$6 (vs. 5 minutes typically for an expert developer). The generated code, for some workloads, exhibited poor quality with excessive overhead, performing worse than EEVDF. The agent required root access, could crash the system during testing, and lacked fallback mechanisms, which also raises safety concerns. These experiments reveal three critical challenges: \textbf{Performance}, ensuring AI schedulers outperform existing ones; \textbf{Safety}, preventing crashes, lockups, and starvation while minimizing privileges; and \textbf{Efficiency}, reducing the 33-minute, \$6 generation cost for practical deployment.

\section{The \sys\ Framework Design and Implementation}
\label{sec:schedcp_framework}

\begin{figure}
    \centering
    \includegraphics[width=\columnwidth]{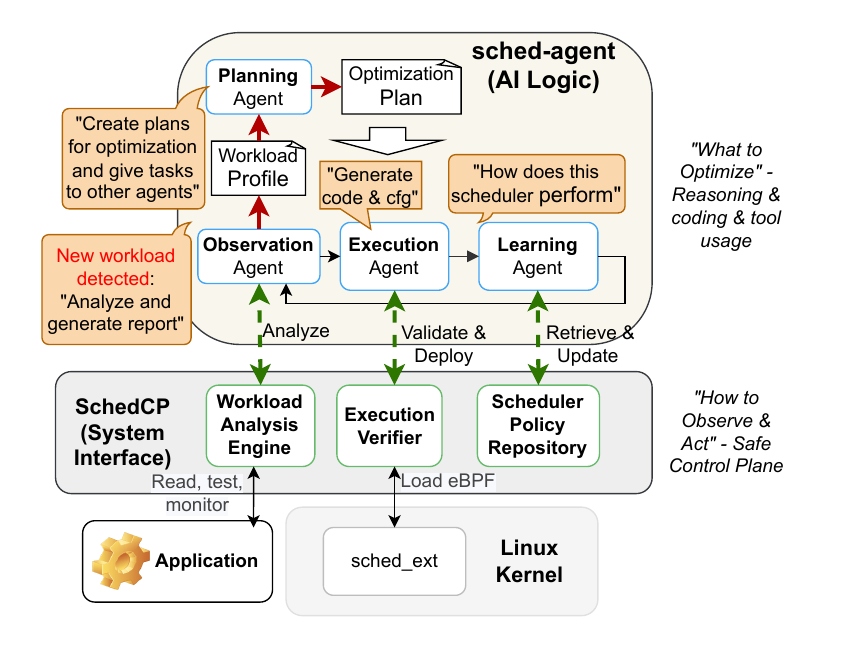}
    \caption{
        \textbf{Overall architecture of \sys\ and \agent.} 
        \sys\ (bottom) provides the system interface with three core services: Workload Analysis Engine, Scheduler Policy Repository, and Execution Verifier.
        \agent\ (top) implements the AI logic through four specialized agents (Observation, Planning, Execution, Learning) working in a closed loop. Red lines show initialization when detecting new workloads, black arrows show optimization loops, and green arrows indicate tool usage by agents.
    }
    \label{fig:frameworkarch}
\end{figure}

As illustrated in Figure~\ref{fig:frameworkarch}, \sys is a secure control plane acting as an 'API for OS optimization,' separating systems infrastructure from AI logic, distinguishing ``what to optimize'' (AI's domain) from ``how to observe and act'' (system's domain). Four key principles govern its design. First, \textbf{decoupling and role separation} ensures future-proofing by treating the AI agent as a performance engineer using a stable set of tools. Second, \textbf{safety-first interface design} addresses the inherent risks of autonomous agents with kernel access by treating AI as potentially non-cautious actors, designing defensive interfaces that prevent catastrophic failures by default, and avoiding granting `root' privileges. Third, \textbf{adaptive context provisioning} addresses LLM agents' constraints from finite context windows and token costs: agents start with minimal summaries and progressively request details as needed. Finally, \textbf{composable tool architecture} follows Unix philosophy by providing atomic tools that let agents construct complex workflows through their reasoning capabilities, enabling novel solution generation rather than constraining them with rigid workflows. Implemented in ~4000 lines of Rust and ~6000 lines of Python (including tests), \sys provides essential tools for any agent to interact with the Linux kernel's scheduler, analogous to how an environment in reinforcement learning provides state, actions, and rewards for learning. It exposes its services via the Model Context Protocol (MCP)~\cite{anthropic2024mcp} through three primary services:

\textbf{1. Workload Analysis Engine} Provides tiered access to system performance data: (1) cost-effective API endpoints with pre-processed summaries (CPU load, memory usage), (2) secure sandbox access to file reading, application building, Linux profiling tools (\texttt{perf}, \texttt{top}) and dynamically attachable eBPF probes, (3) feedback channel reporting post-deployment metrics (percentage change in throughput/latency).

\textbf{2. Scheduler Policy Repository.} Database storing executable eBPF scheduler programs with metadata (natural language descriptions, target workloads, historical performance metrics). It provides APIs for semantic search and retrieval, enabling agents to find relevant schedulers or composable code primitives. To support system evolution, it includes endpoints for updating performance metrics and promoting new policies, reducing generation costs by allowing reuse of proven solutions while maintaining a growing library of scheduling strategies.

\textbf{3. Execution Verifier} includes a multi-stage validation pipeline: (1) kernel's eBPF verifier ensures memory safety and termination, (2) scheduler-specific static analysis detects logic flaws (starvation, unfairness) the standard verifier misses, (3) dynamic validation in secure micro-VM tests correctness and performance. Successful validation issues signed deployment tokens for monitored canary deployments with circuit breakers to revert if performance degrades, eliminating \agent's need for root access.

\section{\agent: A Multi-Agent Framework for OS Optimization}
\label{sec:sched_agents}

Building on \sys, we developed \textbf{\agent}, a multi-agent AI framework implementing in-context reinforcement learning (ICRL)\cite{incontextrl} for scheduler optimization. Using Claude Code's subagent architecture\cite{anthropic2024subagents}, \agent\ decomposes optimization into four distinct ICRL stages through specialized AI assistants with customized prompts, tools, and separate context windows\cite{anthropic2024multiagent}. The framework integrates with container orchestrators (Kubernetes, Docker) to automatically trigger optimization when applications deploy, enabling adaptive strategy refinement based on performance feedback without model retraining.

The \textbf{Observation Agent} builds Workload Profiles by querying the Workload Analysis Engine strategically, starting with high-level summaries from process name and commands then requesting deeper profiling (\texttt{perf stat}, \texttt{top}) based on findings, synthesizing data into natural language descriptions and optimization goals while managing cost-precision tradeoffs. For kernel compilation, it produces profiles like ``CPU-intensive parallel compilation with short-lived processes, inter-process dependencies, targeting makespan minimization.'' The \textbf{Planning Agent} transforms profiles into optimization strategies via the Scheduler Policy Repository, following a decision hierarchy: configuring existing schedulers, generating patches, or composing new schedulers from primitives. The \textbf{Execution Agent} manages development, validation and deployment by synthesizing code artifacts, submitting to the Execution Verifier, interpreting results to refine code or fix logic issues. The \textbf{Learning Agent} completes the ICRL loop by analyzing deployment outcomes (e.g., 45\% makespan reduction), enabling in-session adaptation and updating the repository with refined metrics, deployment contexts, and documented antipatterns.

\section{Preliminary Evaluation}
\label{sec:evaluation}


We validate \sys's effectiveness through four research questions: configuring existing schedulers (RQ1), generating new schedulers for specific workloads (RQ2), cost and efficiency of scheduler generation (RQ3), and iterative refinement improvements (RQ4). Evaluation uses two machines: 86-core Intel Xeon 6787P with 758GB RAM running Linux 6.14, and 8-core Intel Core Ultra 7 258V with 30GB RAM running Linux 6.13. Agents use Claude Code (Opus 4), testing each case three times and averaging results. All experiments successfully created  working custom scheduler configurations or eBPF programs. Future evaluation requires a complete benchmark.

\begin{figure}[h]
\centering
\begin{minipage}{0.48\textwidth}
    \begin{subfigure}[b]{\textwidth}
        \centering
        \includegraphics[width=\textwidth]{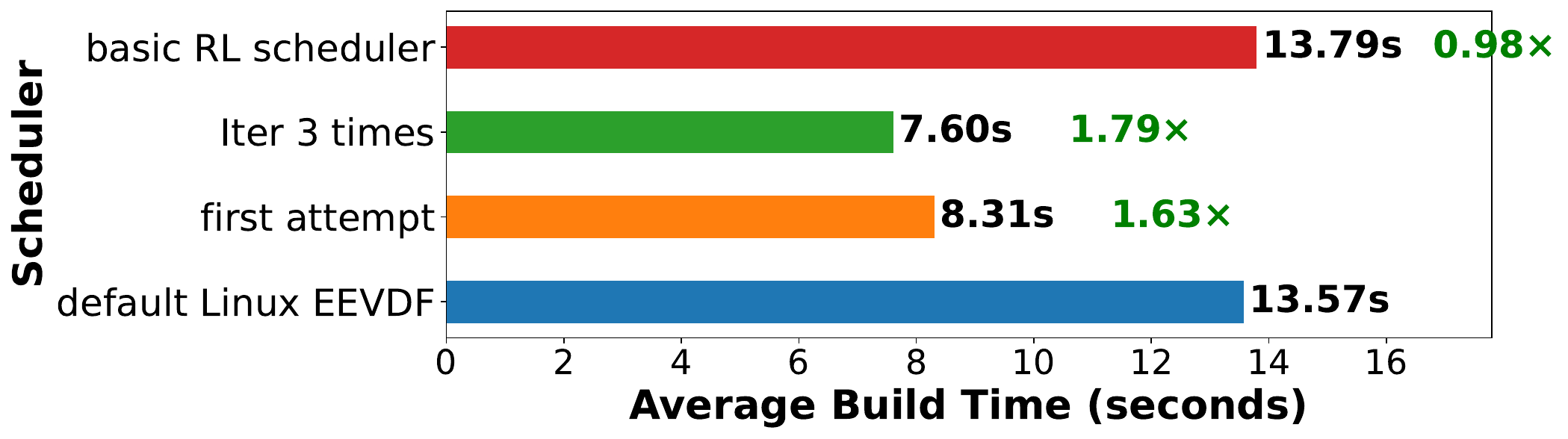}
        \caption{AI configured scheduler for Linux build time}
        \label{fig:performance-comparison}
    \end{subfigure}
    \vspace{0.3cm}
    \begin{subfigure}[b]{\textwidth}
        \centering
        \includegraphics[width=\textwidth]{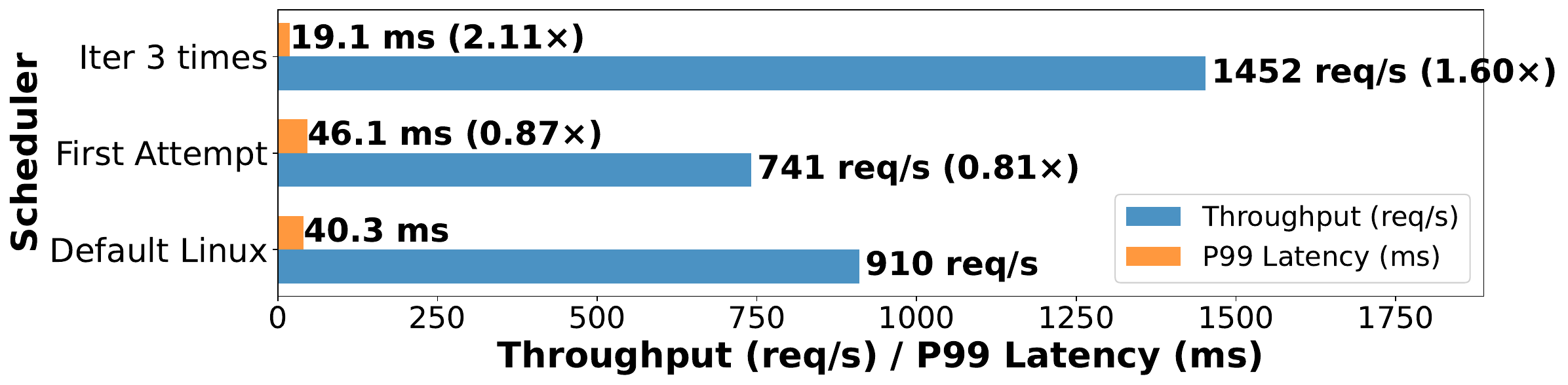}
        \caption{AI configured scheduler for Schbench latency and throughput}
        \label{fig:schbench-comparison}
    \end{subfigure}
\end{minipage}
\hfill
\begin{minipage}{0.48\textwidth}
    \begin{subfigure}[b]{\textwidth}
        \centering
        \includegraphics[width=\textwidth]{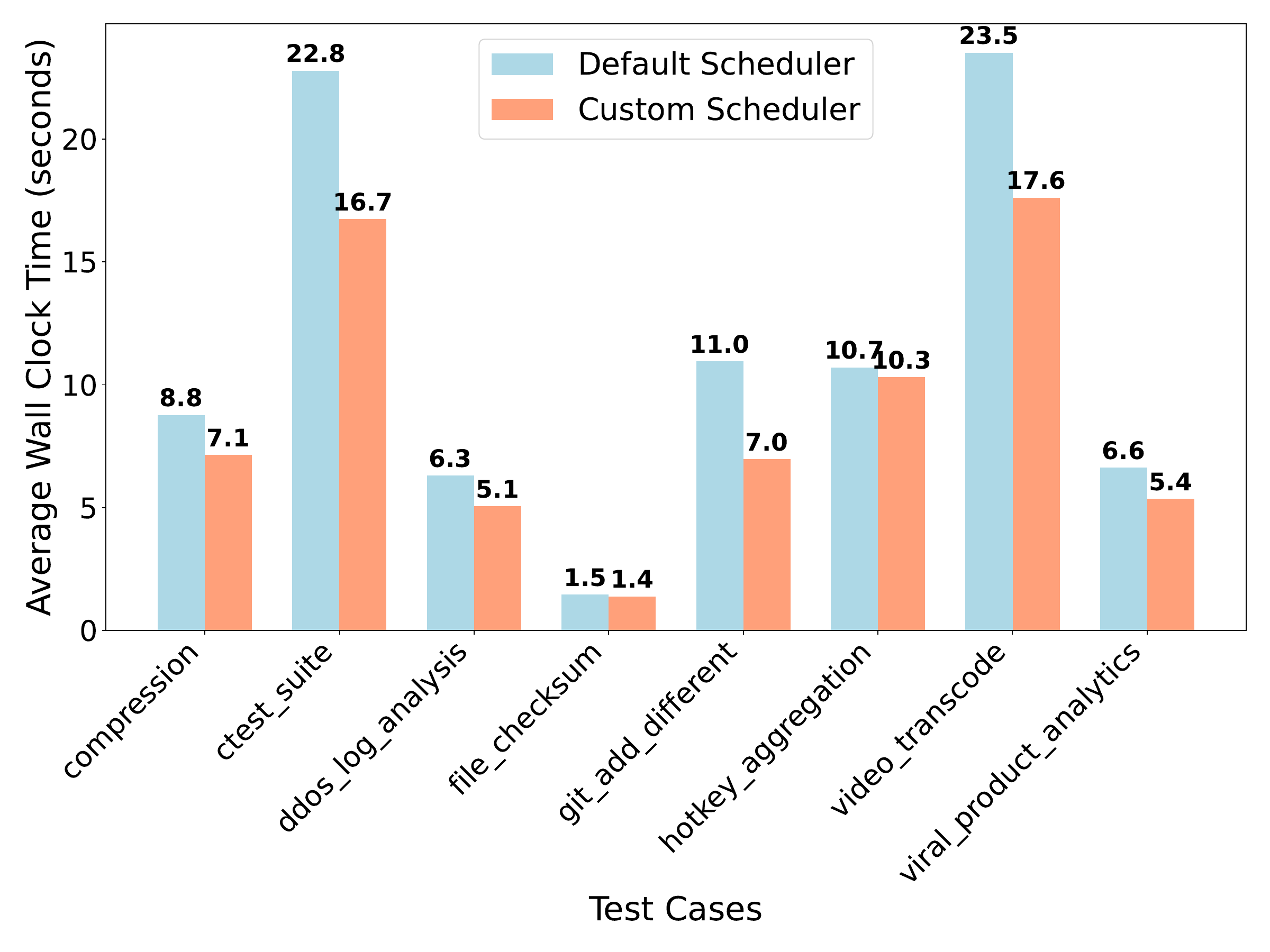}
        \caption{AI-generated scheduler for batch workloads}
        \label{fig:batch-performance}
    \end{subfigure}
\end{minipage}
\caption{Performance evaluation of \sys across different workloads}
\label{fig:combined-performance}
\end{figure}

\textbf{Scheduler Configuration}: For kernel compilation (tinyconfig, ``make -j 172'' on 6.14 source), \sys achieves 1.63× speedup with scx\_rusty initially, then iterative refinement selects scx\_layered for 16\% additional gain, reaching 1.79× total improvement over EEVDF (Figure~\ref{fig:performance-comparison}). Pre-trained RL approaches~\cite{corbet2025ml} show no improvement, likely because they require costly hardware/workload-specific retraining. On schbench~\cite{schbench2016}, initial AI configuration (scx\_bpfland) underperformed, but three refinement iterations identified scx\_rusty as superior: 2.11× better P99 latency and 1.60× higher throughput versus EEVDF (Figure~\ref{fig:schbench-comparison}), demonstrating effective learning from feedback.

\textbf{New Scheduler Synthesis}: For 8 diverse batch workloads (file compression, video transcoding, software testing, data analytics) with long-tail distributions (40 parallel tasks: 39 short, one long), \agent correctly identified the optimization goal and workload pattern, implementing Longest Job First (LJF) scheduling to achieve 20\% average latency reduction (Figure~\ref{fig:batch-performance}). Claude Opus successfully classified all 8 workloads at \$0.15 per analysis, while Claude Sonnet failed. Generation efficiency improved 13× (to 2.5 minutes) with \$0.45 synthesis cost per workload, demonstrating economic viability alongside performance gains.



\bibliographystyle{plain}
\bibliography{sample-base}

\end{document}